\documentclass[11pt,a4paper]{article}
\usepackage[hyperref]{naaclhlt2019}
\usepackage{times}
\usepackage{latexsym}
\usepackage{times}  
\usepackage{helvet}  
\usepackage{courier}  
\usepackage{url}  
\usepackage{graphicx}  
\usepackage{amsmath}
\usepackage{mathtools, nccmath}
\usepackage{subfigure}
\usepackage{times}
\usepackage{booktabs}
\usepackage{indentfirst}
\usepackage{url}

\aclfinalcopy 

\title{Interpretable Neural Embedding with Sparse Words Self-Representation}

\author{Minxue Xia \\
  Vector Lab \\
  JD Finance \\
  \And
  Hao Zhu \\
  Vector Lab \\
  JD Finance \\
  \\}

\date{}

\begin{document}
\maketitle
\begin{abstract}
%
%
%
%

Interpretability benefits the theoretical understanding of representations.
Existing word embeddings are generally dense representations. Hence, the meaning of latent dimensions is difficult to interpret.
This makes word embeddings like a black-box and prevents them from being human-readable and further manipulation.
Many methods employ sparse representation to learn interpretable word embeddings for better interpretability.
However, they also suffer from the unstable issue of grouped selection in $\ell1$ and online dictionary learning. Therefore, they tend to yield different results each time.
To alleviate this challenge, we propose a novel method to associate data self-representation with a shallow neural network to learn expressive, interpretable word embeddings.
In experiments, we report that the resulting word embeddings achieve comparable and even slightly better interpretability than baseline embeddings.
Besides, we also evaluate that our approach performs competitively well on all downstream tasks and outperforms benchmark embeddings on a majority of them. 
\end{abstract}

\section{Introduction}
Word embeddings have gained immense success and obtained state-of-the-art results in a wide range of NLP tasks, such as parsing \cite{intro1,Bansal}, named entity recognition \cite{Guo}, and sentiment analysis \cite{Socher}. One of the main attraction in these embeddings is that they capture the underlying complexity and latent trends in the data. However, a critical issue is that they are difficult to interpret, hence people are unaware of what each dimension represents.
\cite{Spine} examined top-ranked words per dimension ("top dimensions words") in benchmark embeddings like Glove~\cite{Glove} and Word2Vec~\cite{word2vec} and demonstrated that it does not form a semantically coherent group .
For instance, the word 'Internet' is mainly composed by the following top dimensions words in GloVe (as shown in Table.~\ref{tab:example}) which is far from being interpretable.

People have utilized sparse coding that transforms word embeddings into sparse vectors~\cite{spowv,Spine}.
It learns interpretable embeddings by applying non-negative and sparsity constraints on original word embeddings.

However, there are two technical drawbacks for these solutions.
Firstly, it fails to deal with grouped selections due to the employment of $\ell1$ ( Lasso)~\cite{tibshirani1996regression}. 
More specifically, if there is a group of words among which the pairwise correlations are very high, Lasso tends to select only one word from that group and does not care which one is selected~\cite{zou2005regularization}. For instance, there exists two groups of words representing two semantic concepts $college$ and $creature$. Given the target word $youtube$, Lasso tends to select the most similar word in both groups and combine them to represent the target word. Secondly, online learning makes the learned dictionary unstable and unrepeatable. It may yield different basis of the dictionary if we trained the model multiple times. Such phenomenon is observed in SPINE~\cite{Spine}, we examined the overlap between top 5 dimensions only accounts for 7\% when SPINE trained twice.  
Last but not least, rare words are difficult to be considered as the basis of the dictionary, because it cannot be fitted well by other words and cannot fits well to other words.

To alleviate these problems, we propose a method based on words self-representation (SR) to generate highly efficient and interpretable word embeddings. Motivated by the linear algebraic structure of word senses~\cite{arora2018linear}, we extend the linear assertion and assume a word in a subspace can always be expressed as a linear combination of other words in that subspace. Therefore, we use a neural model with Sparse Words Self-Representation (SWSR) to obtain desired embeddings. The self-expressiveness of data guarantees the stability of the basis in the dictionary, and the neural approach can efficiently avoid the issue of grouped selection.

Our contributions can be arranged as follows: 1) We formulate the word embedding with self-expressiveness property of data. 2) We propose a neural model to solve words self-representation with non-negative and sparsity constraints. 3) Last but not least, our approach is competitive to existing methods on many downstream tasks. \\



\begin{table}[!ht]
\centering
\begin{tabular}{|c|c|}\hline
   & Top dimensions words \\ \hline
   & housands, residents, palestinian \\ 
'Internet' & intelligence, government, foreign \\
   & nhl, writer, writers, drama \\\hline
\end{tabular}
\caption{\label{tab:example} An example of un-interpretable of word embeddings. It shows top-ranked words per dimension, each row represents words from top dimensions from GloVe for word 'Internet'. }
\end{table}

\section{Related Work}

To learn interpretable word embeddings, current approaches are generally based on sparse coding. There are two major types of inputs: inputs derived from Positive Point-wise Mutual Information (PPMI) and inputs derived from the neural embedding. \cite{brian,online} apply sparse matrix factorization with non-negativity on the PPMI inputs. \cite{Spine,spowv} employ neural inputs such as pre-trained GloVe and Word2Vec embeddings. Comparing \cite{Spine} and \cite{spowv}, the first approach uses neural-based method to constraint sparse and non-negative embeddings, while the second utilizes the convex optimization to transform dense representations into sparse and interpretable. Other method like \cite{parker} rotates word vectors to enhance interpretability. Our method is constructed based on the neural approach and is different from other methods, since it is more expressive than linear methods like matrix factorization or rotations. Next, words in our method are allowed to participate at varying levels in different dimensions naturally.




\section{Methodology}
In order to obtain stable and interpretable embeddings, we introduce the self-expressiveness property of data, stating each word in a union of subspaces can be efficiently represented as a linear combination of other words. 
Normally, such representation is not unique because there are infinite approaches to express a word through a combination of other words. Ideally, a sparse representation of a word corresponds to a combination of few words from its own subspace. 

In the viewpoint of sparse representation, this avoids learning a dictionary in advance and therefore has a more stable 'dictionary'. The sparse representation NEED TO BE IMPLEMENTED. 
We implement words self-representation in a shallow neural network and also modify the loss functions to pose sparsity constraints on parameters.

\subsection{Learning Embedding with Self-Expressiveness Property}
Given a set of word embeddings $\{X\}_{i=1,...,|V|} \in R^{d\times|V|}$, where $|V|$ is the vocabulary size, and $d$ is the number of dimensions in the input word embeddings. We assume that word representations are drawn from multiple linear subspaces $\{S\}_{i=1,..., K}$. Each word in the subspace can be regarded as a linear combination of other words in the same subspace. This property is called self-expressiveness property (SEP) in \cite{rao2008motion}. We can use a single equation, i.e., $\pmb{X=XC}$,  to represent the property, where $C$ is the self-representation coefficient matrix. It has been shown in \cite{rao2008motion} that, under the assumption that the subspaces are independent, by minimizing certain norms of $C$, $C$ is guaranteed to have a block-diagonal structure (up to certain permutations), i.e., $c_{ij}>0$ if and only if point $x_i$ and point $x_j$ lie in the same subspace.
So we can leverage the words self-representation $C$ to construct new words embedding that is interpretable because only similar words belong to the same subspace and thus have similar representations in $C$.

The self-representation coefficient matrix can be obtained by solving the optimization problem:
\begin{equation}
    \min\|C\|_p\;\;s.t.\;\;X=XC,\;diag(C)=0,\;C_{ij}\ge0
    \label{equ:self}
\end{equation}
where $\|\cdot\|_p$ represents an arbitrary matrix norm, and the optional diagonal constraint on $C$ prevents trivial solutions for sparsity inducing norms, such as the $\ell1$-norm. Various norms for $C$ have been proposed in the literature, e.g., the $\ell1$-norm in Sparse Subspace Clustering (SSC) \cite{elhamifar2009sparse}, the nuclear norm in Low Rank Representation (LRR) \cite{liu2013robust} and Low Rank Subspace Clustering (LRSC) \cite{favaro2011closed}, and the Frobenius norm in Least-Squares Regression \cite{lu2012robust} (LSR).


\subsection{Self-Expressive Layer in Shallow Neural Network}
In this paper, we propose a neural model to obtain words self-representation.
Our goal is to train a shallow neural network where its inputs are well-suited to SEP with a parameter $C$. To this end, we introduce a new layer that encodes the notion of self-expressiveness property. To encode words self-representation under the architecture of a neural network, we reformulate the Eq.~\ref{equ:self} to another form defined as:
\useshortskip
\begin{equation}
    \min_C\frac{1}{2}\|X-XC\|_2^2+\lambda\Omega(C)
    \label{equ:ssc}
\end{equation}
\useshortskip
where $\Omega(C)$ is the regularization terms. Typical terms include sparsity-induced norms $\|\cdot\|_1$ \cite{elhamifar2009sparse}, and nuclear norm \cite{favaro2011closed}.

In contrast to the sparse coding or convex optimization approach of solving the parameter $C$ \cite{elhamifar2009sparse,liu2013robust}, we propose a shallow neural network with a sparsity penalty, a non-negative constraint on $C$ to obtain sparse, interpretable embedding. 
Then we can define the forward step in our method like:
\useshortskip
\begin{equation}
    \hat{X} = Xf(C)
\end{equation}
\useshortskip
where $f$ is an appropriate element-wise activation function to ensure that $C$ is non-negative, and $W\in R^{|V|\times|V|}$ is model parameters. In order to produce non-negative values and exact zeros, we use Capped-ReLU~\cite{Spine} as the activation function to map continuous value into [0, 1]. Mathematically, we define the activation like:
\useshortskip
\begin{equation}
\textbf{Capped-ReLU}(x)=\left\{
\begin{array}{rcl}
0, &if & {x \leq 0}\\
x, &if &{0< x < 1}\\
1, &if &{x \geq 1}\\
\end{array} \right.
\end{equation}
\useshortskip
In this setting, given $X$, our shallow neural network is trained to minimize the following loss function.
\useshortskip
\begin{equation}
    \textbf{L} = \textbf{RL} + \lambda_1\textbf{ASL} + \lambda_2\textbf{PSL}
\end{equation}
\useshortskip
where Reconstruction Loss (\textbf{RL}) is an average loss in reconstructing the input representation from the learned representation, the loss function can be defined with a mean square error like:
\useshortskip
\begin{equation}
    \textbf{RL} = \frac{1}{|V|}\sum^{|V|}\|X_i-\hat{X}_i\|_2^2
\end{equation}
\useshortskip
In order to simulate sparsity-induced norm in our neural model, we penalize any deviation of the observed average activation value in $C$ from the desired average activation value of a given hidden unit. We formulate this Average Sparsity Loss (\textbf{ASL}) as follows.
\useshortskip
\begin{equation}
    \textbf{ASL}=\sum_i^n(\max(\frac{\sum_j^nc_{ij}}{n}-\rho,0))
\end{equation}
\useshortskip
where $\rho$ is the the parameter of sparsity ratio.

Not only obtain k-sparse activation values for input, but we also hope to have discriminative representations such as binary codes. Thus a Partial Sparsity Loss (\textbf{PSL}) term \cite{Spine} is used to penalize values that are neither close to 0 nor 1, pushing them close to either 0 or 1. The formulation of PSL is as following:
\useshortskip
\begin{equation}
    \textbf{PSL}=\frac{1}{n}\sum_i^{n}C_i-C_iC_i^T
\end{equation}
\useshortskip
where $C_i$ is the $i$-th row of the parameter $C$.

\section{Experiments}
To enable fair comparisons, experiments are conducted within the framework of SPINE \cite{Spine}. Briefly, we use pre-trained GloVe and Word2Vec embeddings for training and both embeddings are 300 dimensions long. 


We compare the embedding generated by our model (denoted as SWSR) against GloVe, Word2Vec, as well as SPOWV and SPINE. All hyperparameters in SPINE and SPOWV are under authors' recommendations. To test the quality of these embeddings, we use them in the following benchmark downstream classification tasks: 1) News classification on the 20 Newsgroups dataset (Sports/Computers/Religion); 2) Noun Phrase Bracketing \cite{Lazaridou}; 3) Question Classification \cite{li}. For text classification tasks, we use the average of embeddings of the word in a sentence as features. Algorithms, like SVM, Logistic Regression, Random Forests are experimented.

\subsection{Downstream Task Evaluations}
To test the quality of the embeddings, all performances are evaluated through accuracy (Table \ref{tab:data_class_1} and \ref{tab:data_class_2}). It is clear that embeddings generated by our method generally perform competitively well on all benchmark tasks, and do significantly better on a majority of them. 
\begin{table}[!ht]
\setlength{\belowcaptionskip}{-10pt}
\centering
\resizebox{\columnwidth}{!}{
\begin{tabular}{|c|c|c|c|c|c|c|c|c|}\hline
Tasks &GloVe&SPOWV&SPINE&SWSR \\\hline
Sports&95.31&95.80&\textbf{96.42}&91.53\\
Religion&80.62&81.27&79.80&\textbf{81.59}\\
Computers&71.64&78.16&75.35&\textbf{80.77}\\\hline
NPBracket&73.31&69.89&\textbf{73.58}&68.69 \\\hline
Question.&82.80&87.45&88.75&\textbf{90.28}\\\hline
\end{tabular}}
\caption{\label{tab:data_class_1}Accuracy on three downstream tasks using Glove word embeddings.}
\end{table}

\begin{table}[!ht]
\setlength{\belowcaptionskip}{-10pt}
\centering
  \resizebox{\columnwidth}{!}{
\begin{tabular}{|c|c|c|c|c|}\hline
Tasks &Word2Vec & SPOWV & SPINE & SWSR \\\hline
Sports&93.01&\textbf{95.72}&93.96&94.1\\
Religion&80.09&83.77&83.43 & \textbf{84.17} \\
Computers&71.37&76.32&74.26 & \textbf{81.85}\\\hline
NPBracket& \textbf{77.91}&70.34&72.56 & 70.65\\\hline
Question.&89.02&90.50& 92.34 & \textbf{93.49}\\\hline
\end{tabular}}
\caption{\label{tab:data_class_2}Accuracy on three downstream tasks using Word2Vec word embeddings.}
\end{table}

\subsection{Interpretability}
We investigate interpretability of word embeddings on the word intrusion detection task, seeking to measure how coherent each dimension of these vectors are. For a given dimension, we introduce a word set containing top 5 words in that dimension and a noisy word (also called intruder) from the bottom half of the dimension which ranks high (top 10\%) in other dimensions. 

We follow the automatic evaluation metric \textbf{DistRatio} proposed by \cite{l1} to measure the interpretability of word representations. 
Formulations presented as follows:
\useshortskip
\begin{equation}\small
    \textbf{DistRatio}= \frac{1}{d} \sum_i^d\frac{\textbf{InterDist}}{\textbf{IntraDist}}
\end{equation}
\useshortskip
\useshortskip
\begin{equation}\small
    \textbf{InterDist}= \sum_{w_j \in top_k(i)} \sum_{w_k\in top_k(i)\atop w_k \neq w_j} \frac{dist(w_j, w_k)}{k(k-1)}
\end{equation}
\useshortskip
\useshortskip
\begin{equation}\small
    \textbf{IntraDist}= \sum_{w_j \in top_k(i)} \frac{dist(w_j, w_b)}{k}
\end{equation}
\useshortskip
where $top_k (i)$,  $w_{bi}$ denotes top-k words on dimension $i$ and the intruder word for dimension i. We use Euclidean distance to measure $dist(w_i, w_j)$. $IntraDist$ stands for the average distance between top words, while $InterDist$ represents the average distance between top words and the intruder word.


As we can see from Table \ref{tab:interpret}, SWSR achieves comparable performances with previous baseline approaches and even perform the best on pre-trained Word2Vec embeddings. These results confirm our model generates more interpretable word embeddings. We attribute the success of SWSR to the expressiveness of a neural model, and the non-negativity, as well as sparsity, further lead to semantically coherent dimensions. In this way, SWSR constructs expressiveness and interpretable word representations. 
\begin{table}[!ht]\small
\setlength{\belowcaptionskip}{-15pt}
\centering
\begin{tabular}{|c|c|c|}\hline
Model & Sparsity & DistRatio  \\\hline
Glove& 0\%& 0.99\\
SPINE& 85\% & 1.25\\
SPOWV& 90\%& 1.12\\
SWSR & 95\% & \textbf{1.27} \\\hline
Word2Vec& 0\%&0.98 \\
SPINE& 85\%& 1.28\\
SPOWV& 90\%& 1.08 \\
SWSR & 99\% & \textbf{1.34}  \\\hline
\end{tabular}
\caption{\label{tab:interpret}Average performances across all vector models on the word intrusion task, measured with $DistRatio$.}
\end{table}

\section{Conclusion}
We have presented a method that converts word representations derived from any state-of-the-art embedding models into sparse by implementing Self Representation within a shallow Neural Network. Experiments demonstrate that embeddings generated by our method generally perform competitively than currently recognized methods on a diverse set of downstream tasks. Besides, it also achieves comparable and slightly better interpretability. Moreover, the utilization of words self-representation guarantees selecting multiple words from the same top dimensions to construct interpretable representations. The full usage of data points in SR also ensures the repeatable, stable of the dictionary.

\bibliography{naaclhlt2019}
\bibliographystyle{acl_natbib}
\end{document}